\documentclass{article}

   \usepackage[nonatbib,preprint]{neurips_2022}

\usepackage[utf8]{inputenc} 
\usepackage[T1]{fontenc}    
\usepackage{hyperref}       
\usepackage{url}            
\usepackage{booktabs}       
\usepackage{amsfonts}       
\usepackage{nicefrac}       
\usepackage{microtype}      
\usepackage{xcolor}         
\usepackage{pgfplots}       
\usepackage{wrapfig}        

\title{Experiences from the MediaEval Predicting Media Memorability Task}

\author{%
  Alba G. Seco de Herrera \\
  University of Essex\\
  Colchester, UK CO4 3SQ \\
  \texttt{alba.garciao@essex.ac.uk} \\
  \And
  Mihai Gabriel Constantin\\
  University Politehnica\\ of Bucharest,
  Romania\\
  \AND
  Claire-Hélène Demarty\\
  InterDigital\\
  France\\
  \And
  Camilo Fosco \\
  Massachusetts Institute of\\ Technology Cambridge,
  USA\\
  \And
  Sebastian Halder \\
  University of Essex\\
  Colchester, UK CO4 3SQ\\
    \And
  Graham Healy \\
  Dublin City University\\
  Ireland\\
    \And
  Bogdan Ionescu\\
  University Politehnica\\ of Bucharest,
  Romania\\
  \And
  Ana Matran-Fernandez \\
  University of Essex\\
  Colchester, UK CO4 3SQ\\
  \And
  Alan F. Smeaton\\
  Dublin City University\\
  Ireland\\
    \And
  Mushfika Sultana \\
  University of Essex\\
  Colchester, UK CO4 3SQ\\
  \And
  Lorin Sweeney\\
  Dublin City University\\
  Ireland\\
}

\begin{document}

\maketitle

\begin{abstract}
The Predicting Media Memorability task in the MediaEval evaluation campaign has been running annually since 2018 and several different tasks and data sets have been used in this time. This has allowed us to compare the performance of many memorability prediction techniques on the same data and in a reproducible way and to refine and improve on those techniques. 
The resources created to compute media memorability are now being used by researchers well beyond the actual evaluation campaign. In this paper we present a summary of the task, including the collective lessons we have learned for the research community.
\end{abstract}

\section{Introduction}
%
The Benchmarking Initiative for Multimedia Evaluation (MediaEval)\footnote{\url{https://multimediaeval.github.io/}} offers challenges related to multimedia analysis, retrieval and exploration. Since 2018, the ``Predicting Media Memorability'' task has been organised annually as part of MediaEval. This task aims to understand what makes a video more, or less, memorable.
Several articles describe the overall evolution of the set of sub-tasks and the various approaches that were used to create the resources~\cite{MEoverview2022,MEoverview2021_eeg,MEoverview2021,mediaeval2020memory,MEoverview2019,MEoverview2018,kiziltepe2021annotated,cohendet2019videomem}.
%
%
The data sets used and the sub-tasks in the Predicting Media Memorability task have evolved over the years and are described in Sections~\ref{sec:subtasks}--\ref{sc:data}. An overview of the main techniques that achieved best results is given in Section~\ref{sc:techniques}. The main lessons learned are
described in Section~\ref{sc:lessons} and conclusions are given in Section~\ref{sc:conclusions}.

\section{Sub-tasks over the years}\label{sec:subtasks}
\begin{table}[htb!]
  \caption{Overview of the  sub-tasks performed over the years.}
  \label{tab:tasks}
  \centering
  \begin{tabular}{lccccc}
    \toprule
    \cmidrule(r){1-2}
    Sub-task      & 2018  & 2019 & 2020 & 2021 & 2022 \\
    \midrule
    Short-term video-based prediction & \checkmark  & \checkmark & \checkmark & \checkmark & \checkmark \\
    Long-term video-based prediction & \checkmark  & \checkmark & \checkmark & \checkmark &  \\
    Generalisation &   &  &  & \checkmark & \checkmark \\
    EEG-based prediction &   &  &  & \checkmark & \checkmark \\
    \bottomrule
  \end{tabular}
\end{table}
%
Table~\ref{tab:tasks} summarises the sub-tasks which have run each year in  MediaEval.
The \emph{video-based prediction} sub-tasks require participants to automatically predict memorability scores for videos, i.e., the probability for a video to be remembered in the short- or long-term. 
The \emph{generalisation} sub-task aims to test if the prediction approaches can be generalised to other sources of video data. Specifically, participants are asked to train their systems on one dataset and test them on another dataset. 
Finally, the  \emph{electroencephalogram (EEG) based prediction} sub-task involves the use of EEG signals from users who watched the videos, to predict whether a viewer will remember a video or not (with or without augmenting the EEG data with features extracted from the videos).
Task participants are provided with an extensive data set of videos with memorability annotations, related information, pre-extracted state-of-the-art visual features, and EEG recordings (see Section~\ref{sc:data}).
%

\begin{figure}[!htb]
\centering 
\includegraphics[width=0.65\linewidth]{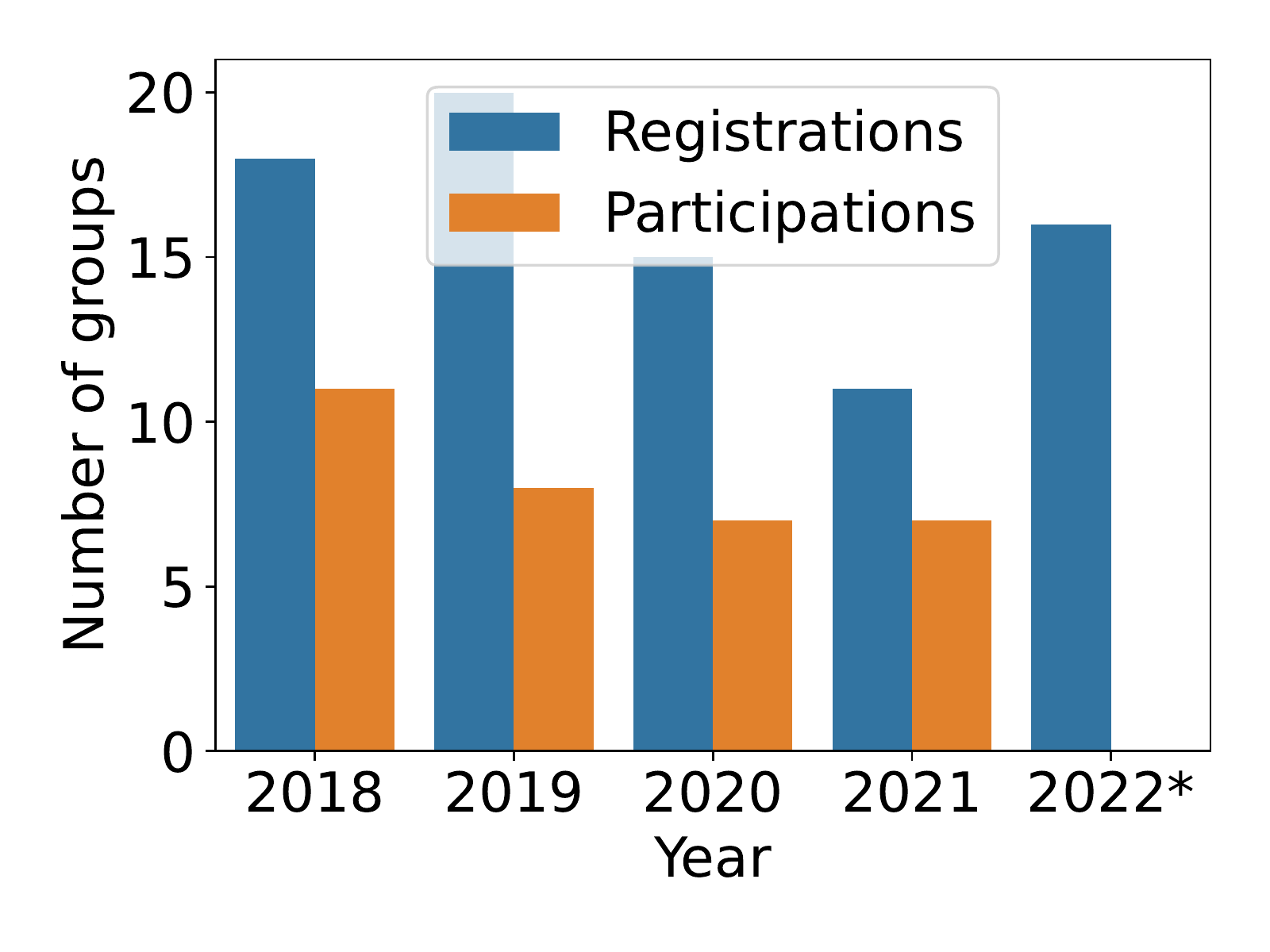}
\caption{Annual participation in the MediaEval Predicting Media Memorability task. Numbers for 2022 are temporary as the deadline for participation is still open.} 
\label{fg:participation}
\end{figure}

\section{Participation over the years}
%
%
In Figure~\ref{fg:participation} the number of groups that registered for the task and that participated are shown across time. Besides the larger availability of benchmarks and data sets nowadays, the task remains attractive for participants. This shows the impact that building data collections and tasks and sharing these resources with other researchers can have on research challenges like computing memorability.

\section{Data over the years}
\label{sc:data}
%
This section describes the evolution of the data released over the years, followed by a description on how  memorability scores have been calculated. 

\subsection{Data sets}
Three video data sets and two EEG-based data sets have been released as part of the MediaEval Predicting Media Memorability task, including pre-extracted features for each.

\begin{figure}[!htb]
\centering
    \includegraphics[width=0.7\textwidth]{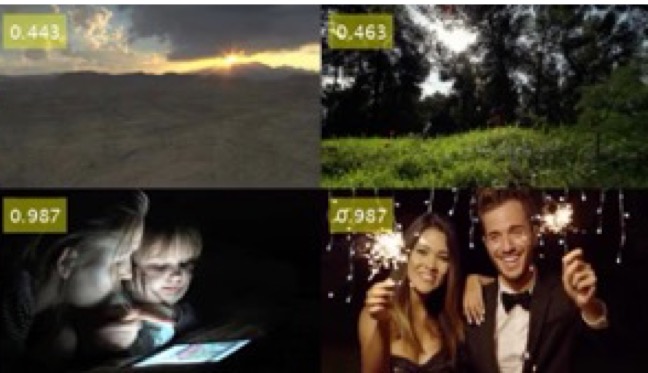}
\caption{A sample of frames (and corresponding memorability scores) from videos in the VideoMem data set.}
\label{fg:videom}
\end{figure}

\paragraph{VideoMem}
~\cite{cohendet2019videomem} is composed of 10,000 7-second long soundless videos shared under a license  allowing their use and redistribution in the context of MediaEval.  They were extracted from high quality cinematic raw stock footage used by professionals when creating content. They are varied and contain different scenes types. 
This data set was used in the 2018, 2019 and 2022 editions of the task. 
In the 2018 and 2019 editions, the data set was split into 80\% for training and development and 20\% for testing. 
In the 2022 edition, 70\% of the videos were released as part of the training set, 15\%  for validation and 15\% for testing.
Figure~\ref{fg:videom} shows a sample of frames from some of the videos in this data set and their corresponding memorability scores.

\paragraph{TRECVid 2019 Video-to-Text}
\cite{ABC2019} contains 6,000 videos with sound, shared under Creative Commons licenses that allow their redistribution. This data  was used in the 2020 and 2021 MediaEval Predicting Media Memorability tasks and was divided  into three sets.
In the 2020 edition, the training set contained 590 videos, the development set 410,  and the test set 500. In the 2021 edition, the training set contained 588 videos, the development set 1,116,  and the test set 500.
The data  contains Twitter Vine videos where various actions are performed. These videos have much more action happening compared to those in the \emph{Memorability 2018} data set, and thus they correspond to more generic use cases. 
Each video consists of a coherent unit in terms of meaning.

\paragraph{Memento10K}
%
\cite{Memento10K}  contains 10,000 3-second long videos depicting in-the-wild scenes, with their associated short-term memorability scores, memorability decay values, action labels, and 5 accompanying captions. 
7,000 videos were released as a training set and 1,500 were provided for validation. The last 1,500 videos were used as the test set for scoring submissions. This data set was used in the 2021 and 2022 editions of the MediaEval Predicting Media Memorability task.

\paragraph{2021 EEG data set}
%
\cite{MEoverview2021_eeg} was collected as part of a pilot study to check the feasibility of finding neural correlates of memory encoding in a short-term memorability task. The videos belong to the TRECVid data set. Both encoding and decoding were recorded in a short-term memorability experiment, but it resulted in a highly imbalanced data set in which the majority of the video clips were correctly remembered, limiting possible analyses and prompting a refinement in the protocol for the 2022 edition of the EEG task.
\paragraph{EEGMem}
\cite{MEoverview2022} comprises pre-extracted features from EEG recordings for 12~subjects captured while they watched a subset of the Memento10k~\cite{Memento10K} videos. Volunteers watched the same videos a second time through a custom-built online portal between 24--72~hours after the video-EEG recording session, where they were required to indicate for each video whether or not they recognised it, providing binary ground truth annotations of memory encoding for long-term memorability. The EEG subtask prompts participants to predict whether a given participant will remember a specific video based on their EEG signals, and includes a cross-subject/transfer learning prompt with data from three subjects being left out of the training set.

\subsection{Measuring video memorability}
%
Two slightly different protocols were used to collect memorability scores depending on the data set~\cite{cohendet2019videomem,Memento10K}. In both cases, participants watched a sequence of videos and  were asked to press the space bar when they recall seeing  a repeated video. For the long-term memorability, participants viewed a new sequence of videos after 24 to 72 hours.

For the VideoMem collection, the memorability scores  were computed from 9,402 participants for short-term, and 3,246 participants for long-term memorability. On average,  38 (at least 30) annotations were collected for each video in the short-term task, whereas in the long-term task, a mean of 13 annotations (at least 9) were collected. 
%
For the TRECVid 2019 Video-to-Text data set, a minimum of 14 annotations per video in the short-term memorability step were collected with a mean of 22. For long-term memorability a minimum of 3 and an average of 7 annotations/video were collected.
For the Memento10K only short-term memorability annotations were acquired. 
The memorability scores were computed with 90 annotations/video on average.

\subsubsection{Memorability scores over the years}
%
%
In the first four editions of the task, each memorability score was calculated as the percentage of correct recognitions among participants (termed \emph{raw scores}). 
In the 2018 and 2019 editions, a linear correction was applied to take into account the different memory retention durations for the short-term scores. The reason for this was that each video repetition happened after different time intervals (between 45--100 videos after the first occurrence) and it is well known that memory decreases linearly with the increase in retention duration. This normalisation aimed to maximise the representativeness of the memory performance after the maximal interval of 100 videos. 
This linear correction was not applied for long-term scores as no specific relationship was observed between the retention durations and the raw scores. The normalisation was not used after the 2020 edition.

In 2021, a second memorability score was introduced based on a different normalisation. This normalised score was based on the assumption that the hit rate for recollection should decay linearly over short timeframes. Also, each video's decay rate may be different, and this is taken into account by using an alpha parameter. For more details regarding this process see~\cite{Memento10K}. In the last edition of the task, in 2022, this normalisation was established.
%

\section{Techniques used}
\label{sc:techniques}
%
Details regarding the methods employed by participants and the results from those methods can be found in the
proceedings of the MediaEval workshop\footnote{\url{http://ceur-ws.org/}}.
With over 30 published papers during the 4 completed editions of the Predicting Media Memorability task and several different methods used by each participating team, some general trends can be noticed with regards to the methods employed by participants.

We observe that fusion methods (whether they process the same modality with more than one system or make use of multi-modal information), and in particular late fusion methods, tend to lead to the top-performing results amongst our participants, regardless of the scale of memorability (short- or long-term) or data set analysed. We conclude that this may be due to better modality representation when creating individual systems for each modality, or even when analysing different aspects of the same modality. This would allow single end-to-end systems to better contribute to a final fusion scheme, and to create better fusion schemes.

Additionally, some participating teams showed an interest in exploring memorability by analysing and predicting correlated concepts. This is an useful approach when studying subjective concepts like memorability, and in the current literature some authors center their study around finding the positive or negative correlation between memorability and other concepts like interestingness or aesthetic appeal~\cite{gygli2013interestingness}. While the results are varied and this type of approach was not always successful, we still consider these types of studies of great value to understand what makes an audiovisual clip more or less memorable. Some examples in this case would be represented by the study of perplexity~\cite{reboud2021exploring}, exploring video semantics~\cite{kleinlein2021thau} or using aesthetic-based features~\cite{constantin2019using}.

Finally, it is encouraging to see that over time there has been an increase of performance of the methods developed over the editions of this competition. We can observe this by analysing the results on VideoMem and TRECVid, which ran for consecutive editions from 2018 to 2021. For example, for the VideoMem data, used during the 2018 and 2019 editions of the task, all participating teams from 2019 scored above the average performance from 2018. Furthermore, two teams had better performance than the top performance from the previous year, in both short- and long-term memorability. A similar trend can be noticed for the TRECVid data, when all teams from 2021 had better results compared with the average from 2020 for short-term memorability, and only one team being behind the average for long-term memorability. Regarding the top results, three teams outperformed the top results from the previous year for short-term and one team for long-term memorability. We will continue monitoring these trends for the upcoming 2022 edition and beyond.

\section{Lessons Learned}
\label{sc:lessons}

In the first two editions of the task, VideoMem was used. This data set contains different scene types, but the videos have limited motion. We found that including motion information is crucial when studying videos compared to only studying images. Therefore the TRECVid 2019 Video-to-Text data set was used in subsequent editions of the task, since it contains more action-rich
video content than VideoMem. Furthermore, for this data set, sound and pre-computed sound features were also provided to task participants.
While no conclusive findings were established, the results from some participants~\cite{SHS2021,reboud2020predicting} suggest that the audio modality provides a degree of useful information during video memorability prediction.

Independently of the source of data, prediction results are significantly lower for long-term memorability than for short-term. One explanation might be the lower number of annotations for long-term scores, but there are other factors at play as well. Attempts to predict long-term scores from the short-term ones were not conclusive, leading to the conclusion that short-term and long-term memories are induced by different characteristics.

Not only does the source of the data influence the quality of the results; memorability annotation quality has a strong impact too. An analysis of the VideoMem data set shows a bias in the distribution of memorability scores towards higher score values. This might be a clue that the task of recognising videos is rather simple as they contain different kinds of information (i.e., motion, color, objects\ldots). Furthermore, the number of annotations is really relevant. For the VideoMem data set, annotation consistency was computed as explained in~\cite{cohendet2019videomem}. It reaches 0.616 (resp.\ 0.364) for the short-term (resp.\ long-term) task (higher is better). These values correspond to respectively strong and moderate correlations according to the usual Spearman scale of interpretation, validating that the number of annotations for the short-term task is sufficient and that a larger number of annotations for the long-term task would have been beneficial.

The TRECVid 2019 Video-to-Text data set has a lower number of annotations and therefore it is very difficult to achieve a strong correlations in the scores. Due to this, it was excluded from the 2022 edition of the task. Unfortunately, annotation acquisition is costly, tedious, and time–consuming. Instead, the Memento10k collection was released since it contains the largest number of annotations/video (an average of 90), although unfortunately it does not contain long-term annotations and the short-term annotations were created without using the sound of the videos. Therefore, in the interest of clarity, standardisation, and the facilitation of more directed inquiry, in the 2022 edition, the scope of the tasks was narrowed, forgoing  raw and long-term memorability scores in favour of normalised short-term memorability scores.

As for the EEG data sets, an error in the data acquisition protocol in the 2021 edition resulted in a heavily imbalanced data set that did not allow for it to be used to make predictions, as most participants had very high recall for the videos (some of them reaching 100\% recall). However, an analysis of the EEG data using time-frequency analyses showed (by comparing brain responses from first vs.\ second presentation of the videos that were repeated) that there are differences that can be used for successful prediction of memorability, thus motivating the acquisition of EEGMem for 2022.

Finally, providing pre-computed features has favoured the participation of more researchers in the task because it helps approach the challenges from different perspectives giving room for a deeper study of the task. The MediaEval workshop is organised annually and  supports open discussion to plan tasks and also evaluate procedures for the future. It has improved motivation across research groups which has turned into many international groups being involved on the organisation of the task favouring its success.

\section{Conclusions}
\label{sc:conclusions}
%
This paper gives a brief overview of the five years of challenges of the MediaEval Predicting Media Memorability Task. It mainly focuses on overviews of how the data, the sub-tasks and the techniques evolved over the years. We also highlight some lessons learned through the organisation of this benchmark over the years and the support of the organisers and the participants.
Although there have been some significant advances, the prediction of media memorability remains an open scientific challenge.

\begin{ack}
%
Financial support was provided by the University of Essex Faculty of Science and Health Research Innovation and Support Fund and by Science Foundation Ireland under Grant Number SFI/12/RC/2289\_P2, cofunded by the European Regional Development Fund.
Financial support was also provided under project AI4Media, a European Excellence Centre for Media, Society and Democracy, H2020 ICT-48-2020, grant \#951911.
\end{ack}

\bibliographystyle{abbrv} 

\bibliography{references}

\end{document}